\documentclass[10pt,twocolumn,letterpaper]{article}

\usepackage{iccv}
\usepackage{times}
\usepackage{epsfig}
\usepackage{graphicx}
\usepackage{amsmath}
\usepackage{amssymb}

\usepackage{mathtools}
\usepackage{bbm}
\usepackage{algorithm}
\usepackage{algpseudocode}
\usepackage{booktabs} % to use \toprule

\usepackage{float} 
\usepackage{subfigure}
\usepackage{multirow}

\usepackage[numbers,sort&compress]{natbib}

\DeclareMathOperator*{\argmin}{arg\,min}

\usepackage[pagebackref=true,breaklinks=true,letterpaper=true,colorlinks,bookmarks=false]{hyperref}

\iccvfinalcopy % *** Uncomment this line for the final submission

\def\iccvPaperID{xxx} % *** Enter the ICCV Paper ID here

\ificcvfinal\fi

\begin{document}

%%%%%%%%% TITLE
\title{Deep Relational Metric Learning}

\newcommand*\samethanks[1][\value{footnote}]{\footnotemark[#1]}

\author{%
% \vspace{-0.6cm}
Wenzhao Zheng\thanks{Equal contribution.}\ , Borui Zhang\samethanks\ , Jiwen Lu\thanks{Corresponding author.}\ , Jie Zhou\\
{Department of Automation, Tsinghua University, China}\\
{Beijing National Research Center for Information Science and Technology, China}\\
{\tt\small \{zhengwz18, zhang-br21\}@mails.tsinghua.edu.cn; \{lujiwen, jzhou\}@tsinghua.edu.cn}\\
% \vspace{-0.5cm}
}

\maketitle
\ificcvfinal\thispagestyle{empty}\fi

\begin{abstract}
This paper presents a deep relational metric learning (DRML) framework for image clustering and retrieval. Most existing deep metric learning methods learn an embedding space with a general objective of increasing interclass distances and decreasing intraclass distances. However, the conventional losses of metric learning usually suppress intraclass variations which might be helpful to identify samples of unseen classes. To address this problem, we propose to adaptively learn an ensemble of features that characterizes an image from different aspects to model both interclass and intraclass distributions. We further employ a relational module to capture the correlations among each feature in the ensemble and construct a graph to represent an image. We then perform relational inference on the graph to integrate the ensemble and obtain a relation-aware embedding to measure the similarities. Extensive experiments on the widely-used CUB-200-2011, Cars196, and Stanford Online Products datasets demonstrate that our framework improves existing deep metric learning methods and achieves very competitive results.
\footnote{Code: \url{https://github.com/zbr17/DRML}.}

\end{abstract} 

\section{Introduction}
How to effectively measure the similarities among examples has been an important problem for many computer vision tasks. Metric learning aims to learn a distance metric under which samples from one class are close to each other and far away from samples from the other classes. Taking advantage of the deep learning technique~\cite{krizhevsky2012imagenet, simonyan2014very, szegedy2015going, he2016deep}, deep metric learning (DML) methods utilize the combination of a convolutional neural network (CNN) and fully connected layers (FCs) to construct a mapping from the image space to an embedding space and employ the Euclidean distance in this space to measure the similarities between samples.
Deep metric learning has outperformed conventional methods by a large margin and demonstrated promising results in a variety of tasks, such as image retrieval~\cite{song2016deep,sohn2016improved,opitz2017bier,zheng2019hardness}, face recognition~\cite{hu2014discriminative,schroff2015facenet,liu2017sphereface,wang2018cosface}, and person re-identification~\cite{shi2016embedding, wang2016joint, zhou2017efficient, chen2017beyond, chen2019deep}.

\begin{figure}[t]
\centering
\includegraphics[width=0.475\textwidth]{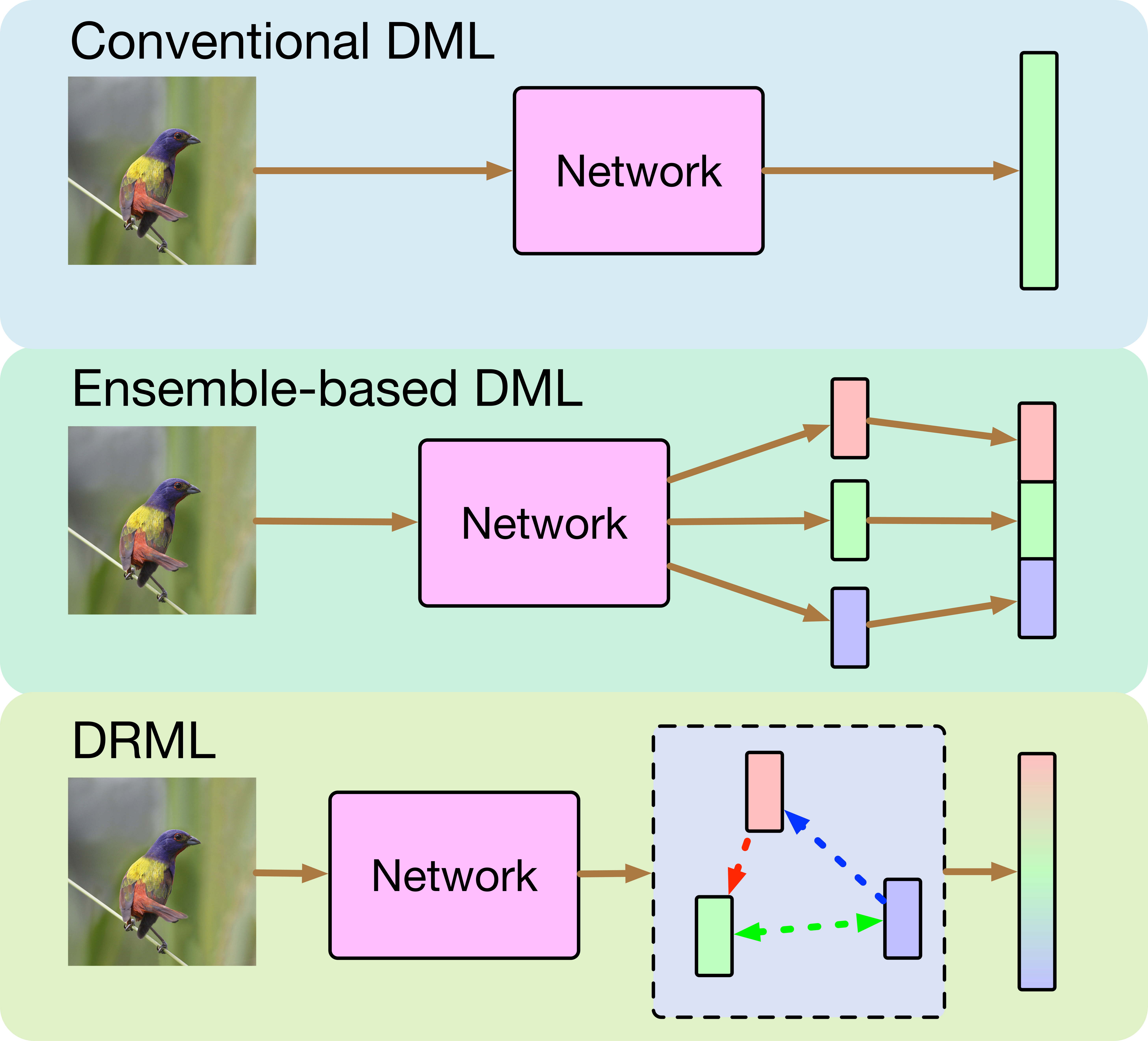}
\caption{Comparisons of the proposed DRML framework with conventional deep metric learning methods and ensemble-based deep metric learning methods. Conventional DML employs a deep neural network to obtain an embedding for each image and measures the similarity of two samples by the Euclidean distance between their corresponding embeddings. Ensemble-based DML learns an ensemble of embeddings but simply concatenates them to obtain the final embedding, which ignores the structural relations among them. Differently, the proposed DRML framework represents one image using an ensemble of features as well as their interactions and then incorporates the relations to infer the final embedding.
(Best viewed in color.)}
\label{fig: motivation}
\vspace{-8mm}
\end{figure}

Most existing metric learning methods utilize a discriminative objective to learn the embedding space, which encourages the metric to discard intraclass variations so that intraclass samples will form a compact cluster with a large margin from the other clusters. The large margin nature of metric learning has been proven useful to improve its robustness and generalization for \emph{seen} class retrieval or classification~\cite{weinberger2009distance, do2012metric, wang2018cosface}. 
However, the essence of metric learning is the ability to generalize to \emph{unseen} classes in the test phase, as is usually evaluated in standard deep metric learning settings~\cite{song2016deep,sohn2016improved,opitz2017bier}. Directly learning a discriminative representation actually harms the performance for unseen class retrieval~\cite{sablayrolles2017should, xuan2020improved}, since the discarded intraclass variations may be useful to differentiate the unseen classes. 
Moreover,
it seems in conflict with the general discriminative objective of metric learning to preserve certain characteristics that is unhelpful for distinguishing training data but might be helpful for unseen test data.
A natural question is raised: \emph{can we learn a discriminative metric that also generalizes well?}

In this paper, we provide a positive answer to this question. We propose a deep relational metric learning (DRML) framework which learns to comprehensively characterize an instance as well as distinguish different instances. We first learn a set of feature extractors to produce an ensemble of features, where each of them describes an image in one aspect. 
We adopt a bottleneck architecture to determine the dominant characteristics of each image and only use samples with the corresponding dominant characteristics to train each feature.
The learned ensemble of features models both interclass differences and intraclass variations, and thus is not discriminative enough to be directly used to compute the distance between images. To effectively measure the similarity between two ensembles of features, we further propose a relational model to discover structural patterns in the feature ensemble and exploit them to obtain a relation-aware embedding. 
The proposed DRML framework can be trained effectively in an end-to-end manner and enjoys the advantage of efficient retrieval similar to existing deep metric learning methods. Differently, our framework induces a stronger relational bias than the combination of convolutional layers and fully connected layers and thus can generalize better to unseen classes.
Figure~\ref{fig: motivation} compares our framework with existing deep metric learning methods.
We conduct extensive experiments on three widely-used datasets which demonstrate the effectiveness of the proposed framework.

\section{Related Work}
\textbf{Deep Metric Learning:}
Deep metric learning methods employ deep neural networks to map an image to an embedding space so that we can effectively measure the similarities between two samples using the Euclidean distance. 
To achieve this, a variety of methods impose a discriminative constraint on the image embeddings~\cite{schroff2015facenet,song2016deep,sohn2016improved,wang2017deep,wang2019multi,yu2019deep,cakir2019deep,ghosh2019learning,sun2020circle,elezi2020group}. For example, the triplet loss~\cite{wang2014learning, schroff2015facenet, cheng2016person} require the distance between each negative pair to be larger than that between each positive pair with a fixed margin. Song~\emph{et al.}~\cite{song2016deep} designed a lifted structured loss to consider all pairs in one batch and implicitly assign larger weights to harder samples. The large number of pairs (or tuples) makes sampling an important component in deep metric learning. A widely used technique is the hard mining strategy which proposes to select false positive tuples for effective training~\cite{schroff2015facenet, huang2016local, yuan2017hard, harwood2017smart, ge2018deep}. In addition, some methods explored other sampling schemes to boost the performance~\cite{wu2017sampling, movshovitz2017no, duan2018deep, ge2018deep, lin2018deep, zhao2018adversarial,zheng2019hardness,duan2019deep,lu2019sampling,suh2019stochastic,roth2020pads}.
For example, Wu~\emph{et al.}~\cite{wu2017sampling} proposed to select pairs with possibilities based on their distances to achieve uniform sampling.

Recently proposed methods begin to consider learning an ensemble of embeddings and concatenate them as the final representation for distance measure~\cite{yuan2017hard,opitz2017bier,kim2018attention,xuan2018deep,sanakoyeu2019divide,milbich2020diva}. As pointed out by \cite{breiman2001random}, the diversity of the learned embeddings in the ensemble is crucial to the final performance of the ensemble. Existing methods achieve diversity by initializing the learners diversely~\cite{opitz2017bier}, using a diversity loss~\cite{kim2018attention, roth2019mic}, weighting each sample adaptively for different learners~\cite{opitz2017bier} or using different samples to train different learners~\cite{yuan2017hard,xuan2018deep,sanakoyeu2019divide}.
However, they all use a simple concatenation to aggregate them,  which fail to consider the structural relations between each entity and lead to inferior generalization performance.

\textbf{Relational Inference:}
The recent success of deep learning in computer vision largely relies on the ability of deep convolutional neural networks to effectively represent images. However, it has been demonstrated that CNNs are not effective for non-Euclidean data and structural relation modeling~\cite{battaglia2018relational}. 
As important complements to CNNs, graph neural networks (GNNs)~\cite{kipf2016semi,kipf2018neural,li2019deepgcns,yang2018glomo,garcia2017few,wang2018videos,jae2018tensorize,li2019graph} have drawn a lot of attention and demonstrated great power for relational inference, which achieve promising results in many tasks like transfer learning~\cite{yang2018glomo}, few-shot learning~\cite{garcia2017few}, video understanding~\cite{wang2018videos}, and scene understanding~\cite{jae2018tensorize}. 
For example, Yang~\emph{et al.}~\cite{yang2018glomo} proposed to capture correlations among pixels and transfer a latent relational graph instead of a generic embedding vector. Wang~\emph{et al.}~\cite{wang2018videos} utilized a graph to represent a video in order to model both spatial and temporal relations.

Motivated by recent advances in relational inference, we propose to represent an image using an ensemble of features and employ a relational model to discover structural patterns. We further perform relational inference to aggregate the ensemble and obtain a relation-aware embedding to measure similarities. Different from existing methods, we propose an end-to-end framework to simultaneously extract generic features and infer their relations to construct a discriminative metric with good generalization ability.

\section{Proposed Approach}

In this section, we first introduce our method of adaptive ensemble learning and then present the formulation of the relation-aware embedding.
Lastly, we detail our framework of deep relational metric learning.

\begin{figure*}[t]
\centering
\includegraphics[width=1\textwidth]{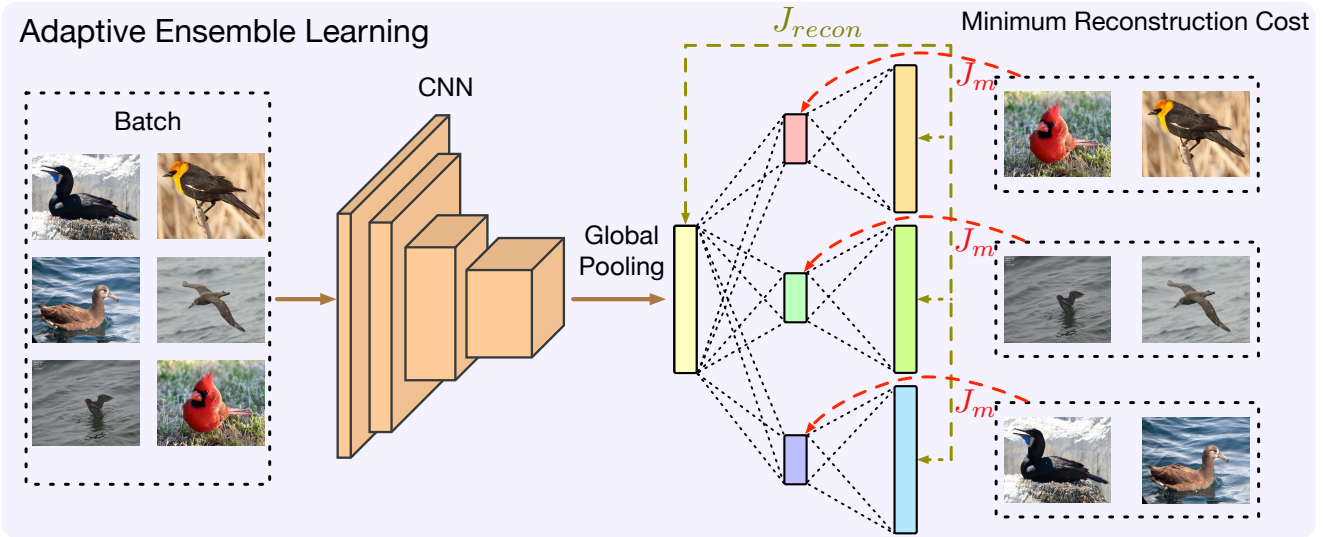}
\caption{Illustration of the proposed adaptive ensemble learning method. We use the output after the global pooling layer of the CNN as a global feature and employ $K$ parallel  fully connected layers to obtain the ensemble of individual features. We then use $K$ decoders to reconstruct the global feature from the corresponding individual features, where we impose a L2 reconstruction loss. We assign each image to the individual feature with the minimum reconstruction cost and only use them to train this branch with a discriminative loss $J_m$ so that each individual feature can describe the image from one aspect. (Best viewed in color.)
} 
\label{fig:ensemble}
\vspace{-5mm}
\end{figure*}

\subsection{Adaptive Ensemble Learning}
We denote a set of images by $\mathbf{X}=\{\mathbf{x}_1, \mathbf{x}_2, \cdots , \mathbf{x}_N\}$ with $L = \{l_1, l_2, \cdots, l_N \}$ as their corresponding labels, where each $l_i \in \{1,2, \cdots, n \}$ means that $\mathbf{x}_i$ is from the $l_i$th class. Deep metric learning employs a deep neural network to learn a mapping $\mathbf{e}_\theta : \pmb{\mathcal{X}} \xrightarrow{e_\theta} \mathbb{R}^D$ from the image space $\pmb{\mathcal{X}}$ to an embedding space $\mathbb{R}^D$, and defines the learned metric as the Euclidean distance in the embedding space:
\begin{eqnarray}\label{equ:distance}
D (\mathbf{x}_i, \mathbf{x}_j; \theta) = ||\mathbf{e}_\theta(\mathbf{x}_i) -\mathbf{e}_\theta(\mathbf{x}_j)||_2,
\end{eqnarray}
where $||\cdot ||_2$ denotes the L2 norm and $\theta$ is the parameters of the embedding network. 

The general objective of deep metric learning is to pull closer samples from the same class and push further samples from different classes, which 
progressively discards intraclass variations and enlarges the distance margin between positive and negative pairs to enhance robustness. However, the discarded intraclass variations might contain information that is useful to differentiate unseen classes, and thus harm the generalization ability of the learned metric.

To address the above problem, we propose to learn an ensemble of individual features to comprehensively characterize an image from different aspects. 
The characteristics of an image include class-relevant ones like textures and class-irrelevant ones like backgrounds, resulting in the interclass and intraclass distributions, respectively.
Instead of discarding the intraclass variations, we propose to capture both interclass and intraclass distributions by enforcing different individual features to capture different characteristics. We adaptively learn the feature ensemble and train each individual feature using only a subset of the samples.

\begin{figure*}[t]
\centering
\includegraphics[width=1\textwidth]{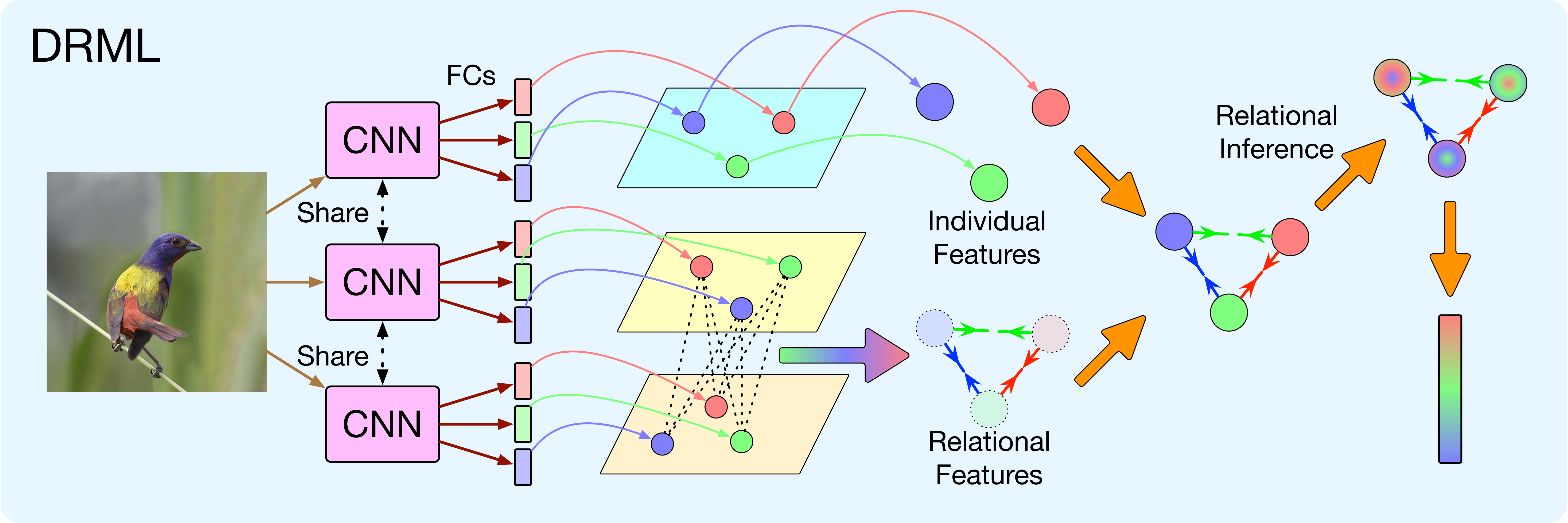}
\caption{The flowchart of the proposed DRML framework. We employ a shared CNN network and $3K$ different fully connected layers to extract $K$ individual features and $2K$ meta-relational features. We use the meta-relational features to produce the relational features between each pair in the ensemble of individual features. We then construct a graph to represent the input image by using the individual features and relational features as the node features and edge features, respectively. We further perform relational inference on the graph to update each individual feature and concatenate them to obtain a relation-aware embedding to measure similarities. 
(Best viewed in color.)} 
\label{fig:framework}
\vspace{-5mm}
\end{figure*}

Formally, we use a trunk CNN to extracts a global feature $\mathbf{y} = \mathbf{f}(\mathbf{x}) \in \mathbb{R}^{d'}$ (i.e., the feature after the final global pooling operation) of an image $\mathbf{x}$ and propose to learn $K$ parallel fully connected layers $ \mathbf{g}_1, \mathbf{g}_2, \cdots, \mathbf{g}_K $ to obtain an ensemble of individual features $\mathbf{G}(\mathbf{x}) = \{ \mathbf{g}_1(\mathbf{f}(\mathbf{x})), \mathbf{g}_2(\mathbf{f}(\mathbf{x})), \cdots, \mathbf{g}_K(\mathbf{f}(\mathbf{x})) \}$, where each $\mathbf{g}_i(\mathbf{f}(\mathbf{x})) \in \mathbb{R}^d (d<d')$ is expected to characterize one aspect of the image. In this paper, we use the set of feature extractors $\mathbf{G}$ to represent both the trunk CNN and the subsequent parallel FCs for convenience. 

To achieve this, we employ a bottleneck structure to determine the dominant characteristics of each image and only train each individual feature using a subset of training samples with the corresponding dominant characteristics.  
Specifically, we use a set of $K$ fully connected layers $\mathbf{P} = \{ \mathbf{p}_1, \mathbf{p}_2, \cdots, \mathbf{p}_K \}$ as decoders to map each individual feature back to the $d'$-dimension space $\mathbb{R}^{d'}$ to reconstruct the global feature $\mathbf{y}$. We adopt the L2 loss the train the reconstruction layers $\mathbf{P}$:
\begin{eqnarray}\label{equ:recon}
\min_{\theta_{\mathbf{P}}} J_{recon} (\mathbf{y}, \mathbf{P}(\mathbf{G}(\mathbf{x}))) = \min_{\theta_{\mathbf{P}}} \frac{1}{K} \sum_{k=1}^{K} ||\mathbf{p}_k(\mathbf{g}_k(\mathbf{y})) - \mathbf{y} ||_2, 
\end{eqnarray}
where $||\cdot ||_2$ denotes the L2 norm. Note that we only use $J_{recon}$ to learn the parameters of the decoders which has no effect on the individual features.

We assume that different images can possess different sets of dominant characteristics that can best describe this image. For example, color is an important characteristic to distinguish different bird species, but it usually cannot be used to identify a specific car model since one car model can present different colors.
We further argue that if one individual feature mainly represents the dominant characteristics of the image, it should contain more information and we should be able to better reconstruct the global feature. 
Therefore, we first classify each image to the individual feature with the minimum reconstruction cost:
\begin{eqnarray}\label{equ:classify}
 k^*(\mathbf{x}) = \argmin_{k} ||\mathbf{p}_k(\mathbf{g}_k(\mathbf{y})) - \mathbf{y} ||_2.
\end{eqnarray}

We then use the corresponding classified subset of images to train each individual feature:
\begin{eqnarray}\label{equ:individual}
\min_{\theta_{\mathbf{G}_k}} J_{ind}^k (\{\mathbf{x}\}) = \min_{\theta_{\mathbf{G}_k}} \sum_{\{\mathbf{x} | k^*(\mathbf{x})=k \} } J_m (\{\mathbf{x}\}),
\end{eqnarray}
where $\theta_{\mathbf{G}_k}$ represents the parameters of the trunk CNN $\mathbf{f}$ and the $k$th fully connected layer $\mathbf{g}_k$, and $J_m$ can be any discriminative loss of metric learning like the triplet loss~\cite{schroff2015facenet} or Proxy-NCA loss~\cite{movshovitz2017no}, and $\{\mathbf{x}\}$ denotes a tuple depending on the specific form of $J_m$.  

The $K$ decoders (i.e., $\mathbf{P}$) are trained with $J_{recon}$ using all the samples, while each encoder (i.e., $\mathbf{g}$) is trained with $J_{ind}$ using the subset of samples with the minimal reconstruction cost.
The set of decoders serves as a classifier to assign each sample to the feature extractor that best preserves its characteristics. 
Still, we allow the gradients of $J_{ind}$ through the backbone CNN so that it is trained using all the samples.

Intuitively, we learn the feature ensemble to comprehensively represent an image as well as reinforce the characteristics that can well recognize samples. We only use a subset of training samples to learn each individual feature, which essentially imposes a less discriminative constraint. 
Still, $J_{ind}$ restricts the feature ensembles of a positive pair to have at least one feature that can well determine the resemblance between them and strengthens only the dominant characteristics to differentiate the negative pair to avoid pushing away features on which they actually resemble each other.

\subsection{Relation-Aware Embedding Learning}
While the learned feature ensemble can comprehensively represent an image, how to exploit them to measure the similarities between images is not trivial. Simply concatenating the features in the ensemble and employing the Euclidean distance to measure the similarities is not effective since we do not explicitly constrain the distance between the concatenation of individual features. The difference between a positive pair can be large and the difference between a negative pair can be small. However, there exist certain relational patterns other than simple distances in the distributions of the individual features that can distinguish them, since we describe each image from different aspects and preserve the dominant characteristics that best characterize it. 

We hence propose a relational model $\mathbf{h}(\mathbf{G}(\mathbf{x}))$ that can well capture the relations among the individual features and emploit them to aggregate the features to obtain the final embedding. Ideally, we want to model each of the relations between two features using separate functions, i.e. $\mathbf{R} = \{ \mathbf{R}_{ij}, i,j \in {1,2,\cdots, K} \}$, but the total number of functions to be learned is with the complexity of $O(K^2)$, which is prone to overfitting. Instead, we propose to learn two set of meta-relational features $\mathbf{A} = \{ \mathbf{a}_1, \mathbf{a}_2, \cdots, \mathbf{a}_K \}$ and $\mathbf{B} = \{ \mathbf{b}_1, \mathbf{b}_2, \cdots, \mathbf{b}_K \}$ to represent the relations among features. The relation feature between the $i$th and $j$th individual feature of an image $\mathbf{x}$ is formulated by the difference between the corresponding meta-relational features: 
\begin{eqnarray}\label{equ:relation}
\mathbf{R}_{ij}(\mathbf{x}) = \mathbf{a}_i(\mathbf{f}(\mathbf{x})) - \mathbf{b}_j(\mathbf{f}(\mathbf{x})),
\end{eqnarray}
where we only need $2K$ functions to represent all the relations.
We can regard $\mathbf{A}$ and $\mathbf{B}$ as representations of the individual feature ensemble in a relational space, where the difference between two features indicates the relations.  
Note that the relations are not symmetric, i.e., $\mathbf{R}_{ij} \neq \mathbf{R}_{ji}$. This enables the proposed relational model to encode more complex and structural relations. 

Having obtained the relations, we construct a graph with the individual features as node features and the relational features as edge features. The graph representation of an image encodes more information than the conventional embedding representation, as it not only expresses the characteristics of the image but also the relations among them.

We can then employ any Graph Neural Networks to perform the relational inference and obtain a final global graph embedding for each image.
In this work, we design a simple but effective module to aggregate the individual features based on the learned relations. We first produce the messages sent to each individual feature by a weighted sum of all features:
\begin{eqnarray}\label{equ:message}
\mathbf{M}_i = \sum_{j=1}^K r_{ji} \mathbf{g}_{j}(\mathbf{f}(\mathbf{x})).
\end{eqnarray}
$r_{ji}$ is the weight of the $j$th individual feature computed by the normalized relational score:
\begin{eqnarray}\label{equ:weight}
r_{ji} = \frac{s(\mathbf{R}_{ji}(\mathbf{x}))}{\sum_{j=1}^K  s(\mathbf{R}_{ji}(\mathbf{x}))}, 
\end{eqnarray}
where $s(\cdot)$ is a relational score function instantiated by a fully connected layer to compute a score reflecting the ``closeness" between two individual features. 

We then integrate the incoming message into the individual feature and update it as follows:
\begin{eqnarray}\label{equ:aggregate}
\mathbf{g}^u_i(\mathbf{x}) = \mathbf{U}(concat([\mathbf{g}_i(\mathbf{f}(\mathbf{x})); \mathbf{M}_i])),
\end{eqnarray}
where $concat(\cdot)$ denotes the concatenation operation and $\mathbf{U}$ is an updater instantiated by a fully connected layer.

We concatenate all the updated individual features to obtain the final relation-aware embedding of the image $\mathbf{z} = concat([\mathbf{g}^u_1(\mathbf{x}); \mathbf{g}^u_2(\mathbf{x}); \cdots; \mathbf{g}^u_K(\mathbf{x}) ])$ and measure the similarities between two images by the Euclidean distance of the relation-aware embeddings:
\begin{eqnarray}\label{equ:metric_final}
D(\mathbf{x}_i, \mathbf{x}_j) = ||\mathbf{h}(\mathbf{G}(\mathbf{x}_i)) - \mathbf{h}(\mathbf{G}(\mathbf{x}_j))||_2 = ||\mathbf{z}_i - \mathbf{z}_j||_2.
\end{eqnarray}
We then apply an embedding loss $J_{emb}$ to the embeddings similar to existing deep metric learning methods.

Though eventually our model still uses a single embedding to represent the similarity between images, the underlying model to obtain the embedding is essentially different from conventional deep metric learning methods. The proposed DRML framework first extracts an ensemble of features to comprehensively characterize an image and then employs a relational model to exploit their relations to produce the final embedding.
It exploits more structural information and induces a stronger relational bias than the simple combination of CNNs and FCs.

\subsection{Deep Relational Metric Learning}
We present our proposed DRML framework, which is composed of the trunk CNN to extract the comprehensive feature, a set of FCs $\{ \mathbf{g}_1, \mathbf{g}_2, \cdots, \mathbf{g}_K \}$ to produce an ensemble of individual features, and a relational model $\mathbf{h}$ to obtain the relation-aware embedding to compute the distance, as shown in Figure~\ref{fig:framework}.

The feature extractors $\mathbf{G}$ and the relational model $\mathbf{h}$ can be trained simultaneously, and we block the effect of $J_{emb}$ on $\mathbf{G}$. This enables the set of feature extractors $\mathbf{G}$ to focus on learning a more comprehensive representation of the images, while the relational model $\mathbf{h}$ aims to mine the relations among the individual features and discover patterns to identify the images. The two modules work together towards the same direction only with slightly different purposes. 

We use the ensemble loss $J_{ensem}$ to train the set of feature extractors $\mathbf{G}$, the reconstruction loss $J_{recon}$ to learn the decoders $\mathbf{P}$, and the embedding loss $J_{emb}$ to train the relational model $\mathbf{h}$:
\begin{eqnarray}\label{equ:objective}
\min_{\theta_{\mathbf{G}}, \theta_{\mathbf{P}}, \theta_{\mathbf{h}}} J = \min_{\theta_{\mathbf{G}}} J_{ensem} + \lambda_1 \min_{\theta_{\mathbf{P}}} J_{recon} + \lambda_2 \min_{\theta_{\mathbf{h}}} J_{emb},
\end{eqnarray}
where $J_{ensem} = \sum_{k=1}^{K} J_{ind}^k$ represents the ensemble objective, $\theta_{\mathbf{G}}$, $\theta_{\mathbf{P}}$ and $\theta_{\mathbf{h}}$ are the parameters of $\mathbf{G}$, $\mathbf{P}$, and $\mathbf{h}$, respectively, and $\lambda_1$ and $\lambda_2$ are two pre-defined parameters to balance the contributions of the three losses. Though $\mathbf{G}$ has an effect on the final embedding $\mathbf{z}$ and thus $J_{emb}$, we do not backpropagate the gradients of $J_{emb}$ through $\mathbf{G}$.

Our framework can be generally applied to existing deep metric learning methods with various loss functions and sampling strategies. 
In this paper, we adopt the same discriminative loss as the individual feature loss $J_m$ and the embedding loss $J_{emb}$, but note that our DRML framework does not stipulate this identity.
We can use different sampling strategies to sample tuples based on the employed loss and use them to compute the embedding loss $J_{emb}$.

The proposed DRML framework can be trained in an end-to-end manner. We use an ensemble of individual features to describe an image and employ a relational model with stronger relational inductive biases than existing deep metric learning methods to improve the generalization ability.
Still, our framework preserves the advantage of efficient retrieval, where we can pre-compute the embeddings of all the images in the gallery and simply compare the embedding of a query image with them.

\section{Experiments}
In this section, we evaluated our framework on three widely used datasets on image clustering and retrieval tasks. 

\newcommand{\tablesize}{\footnotesize}

\subsection{Datasets}
To evaluate the generalization performance of our framework on unseen classes, we followed existing methods~\cite{song2016deep,zheng2019hardness,sohn2016improved} to conduct experiments under a zero-shot setting, where the training subset has no intersection with the test subset. The three datasets were split as follows.

The \textbf{CUB-200-2011} dataset~\cite{wah2011caltech} includes 200 bird species of 11,788 images. We used the first 100 species of 5,864 images as the training subset and the remaining 100 species of 5,924 images as the test subset. 
The \textbf{Cars196} dataset~\cite{krause20133d} contains 196 car makes and models of 16,185 images. We used the first 98 models of 8,054 images as the training subset and the remaining 100 models of 8,131 images as the test subset. 
The \textbf{Stanford Online Products} dataset~\cite{song2016deep} includes 22,634 products of 120,053 images collected from eBay.com. We
used the first 11,318 products of 59,551 images as the training subset and the remaining 11,316 products of 60,502 images as the test subset.

\subsection{Implementation Details}
We used the PyTorch package in all the following experiments.  
We used the ImageNet~\cite{russakovsky2015imagenet} pretrained CNN as the backbone model in our framework for a fair comparison with most existing deep metric learning methods. We set the number of individual features $K$ to 4. We added twelve randomly initialized fully connected layers after the global pooling layer, where four of them output the individual features in the ensemble and eight of them output the meta-relational features. We set the dimensions of individual features and meta-relation features to 128.
We set the output dimension of the updater to 128 which is the same with that of the individual features.
We first normalized all the images to $256 \times 256$. 
For training, we performed data augmentation by standard random cropping to $227 \times 227$ and random horizontal mirror.
We fixed the batch size to 80 and set the learning rate to $10^{-5}$ for the CNN and $10^{-4}$ for the other fully connected layers. 
We set $\lambda_1$ and $\lambda_2$ to 0.1 and 10 to balance the effects of the losses.

\subsection{Evaluation Metrics}
We conducted experiments on both image clustering and retrieval tasks following existing works~\cite{song2016deep,zheng2019hardness,sohn2016improved}. For the image clustering task, we adopted the normalized mutual information (NMI) and $F_1$ score as the evaluation metrics. NMI computes the mutual information between clusters and the ground truth classes normalized by the arithmetic mean of the entropies of them, i.e., $\text{NMI}(\Omega,\mathbb{C})=\frac{2I(\Omega;\mathbb{C})}{H(\Omega)+H(\mathbb{C})}$,
where $\Omega=\{\omega _1, \cdots, \omega _K \}$ denotes the set of clusters and $\mathbb{C}=\{c_1, \cdots, c_K\}$ denotes the set of ground truth classes. $\omega_i$ includes samples predicted to belong to the $i$th cluster, and $c_j$ includes samples from the $j$th class.
The $F_1$ score computes the harmonic mean of the precision and recall
, i.e., $\text{F}_1=\frac{2PR}{P+R}$. 
For the image retrieval task, we adopted the Recall@Ks as the evaluation metrics. Recall@K measures the percentage of legitimate samples in the test subset. A sample is deemed legitimate if there exists at least one positive sample in its K nearest neighbors.
We direct interesting readers to \cite{song2016deep} for detailed explanation.

\begin{table}[t] \tablesize
\centering
\caption{Effect of different numbers of individual features.}
\label{tab:number_feature}
\vspace{5pt}
\begin{tabular}{cccccc}
\toprule 
No. of features & R@1 & R@2 & R@4  &R@8 &NMI \\
\midrule  
2   & 68.0 & 77.4  & 85.7 & 90.9 & 68.6 \\
4  & \textbf{68.7}  & \textbf{78.6} & \textbf{86.3}  & \textbf{91.6} & \textbf{69.3} \\
8   & 66.1  & 76.0 & 84.8  & 90.9  & 68.0  \\
16  & 65.7  & 76.2 & 84.2  & 90.1  & 66.8  \\
\bottomrule
\end{tabular}
\vspace{-3mm}
\end{table}

\begin{table}[t] \tablesize
\centering
\caption{Ablation study of different model settings.}
\label{tab:ablation}
\vspace{5pt}
\begin{tabular}{lccccc}
\toprule 
Methods & R@1 & R@2 & R@4  &R@8 &NMI \\
\midrule  
DiVA~\cite{milbich2020diva}   & 66.4 & 77.2  & 85.8 & 91.5 & \textbf{69.6} \\
DRML-DiVA   & 63.8  & 74.8 & 84.1  & 90.2  & 67.5  \\
AEL-PA   & 60.5  & 71.8 & 81.1  & 88.1 & 62.6 \\
DRML-PA  & \textbf{68.7}  & \textbf{78.6} & \textbf{86.3}  & \textbf{91.6}  & 69.3  \\
\bottomrule
\end{tabular}
\vspace{-3mm}
\end{table}

%\setlength\tabcolsep{7pt}
%\vspace{-3.4mm}
\begin{table}[t] \tablesize
\centering
\caption{Ablation study using different loss functions.}
\label{ablation_loss}
\vspace{5pt}
\begin{tabular}{lcccccccc}
\toprule
Method & R@1 & R@2 & R@4  & R@8 & NMI \\
\midrule 
DRML w/o $J_{ensem}$ & 45.6  & 59.5  & 72.2  & 82.5  & 56.3  \\
DRML w/o $J_{emb}$ & 60.4  & 71.4  & 81.1  & 88.4  & 61.3  \\
DRML w/o $J_{recon}$ & 66.6  & 77.3  & 85.3  & 90.7  & 69.3  \\
DRML-PA & \textbf{68.8 } & \textbf{79.3 } & \textbf{87.1 } & \textbf{91.6 } & \textbf{71.6 } \\
\bottomrule
\end{tabular}
\vspace{-7mm}
\end{table}

\subsection{Results and Analysis}

\textbf{Number of individual features:}
The proposed DRML framework employs $K$ individual features to describe an image from different aspects. 
We conducted experiments on the CUB-200-2011 dataset to analyze the effect of the number of individual features, as shown in Table~\ref{tab:number_feature}. 
We apply the proposed DRML to the ProxyAnchor loss~\cite{kim2020proxy} (i.e., DRML-PA) and fix the embedding size to 512. 
We use 2, 4, 8, and 16 as the number of individual features, rendering the size of each individual feature to 256, 128, 64, and 32, respectively.
We see that our method achieves the best result at $K=4$, and using a larger $K$ slightly harms the performance. This is because, with a larger $K$, each individual feature possesses a lower information capacity which cannot fully describe the main characteristics of an image.

\textbf{Ablation study of each module:}
We conducted an ablation study to analyze the effectiveness of each module in the proposed DRML framework. 
We first remove the relational module and only employ the proposed adaptive ensemble learning method to learn a set of individual features using the ProxyAnchor loss and simply concatenate them as the final embedding (i.e., AEL-PA). 
We also adopt the state-of-the-art ensemble learning method DiVA~\cite{milbich2020diva} to learn the individual features and employ the proposed relational module to model the relations to infer the final embedding (i.e., DRML-DiVA).
Table~\ref{tab:ablation} shows the experimental results on the CUB-200-2011 dataset using BN-Inception~\cite{ioffe2015batch} as the trunk CNN.
We observe that adding the relational module hinders the performance of DiVA. We suspect this is because the ensemble learned by DiVA cannot describe an image from different aspects and our relational module cannot capture their relations.
On the contrary, though the concatenation of the proposed adaptive ensemble is not discriminative enough to distinguish different samples, further employing a relational module improves the performance dramatically. 
This demonstrates the effectiveness of exploiting relations to infer the final embedding.

\begin{figure}[t] 
\centering
\includegraphics[width=0.33\textwidth]{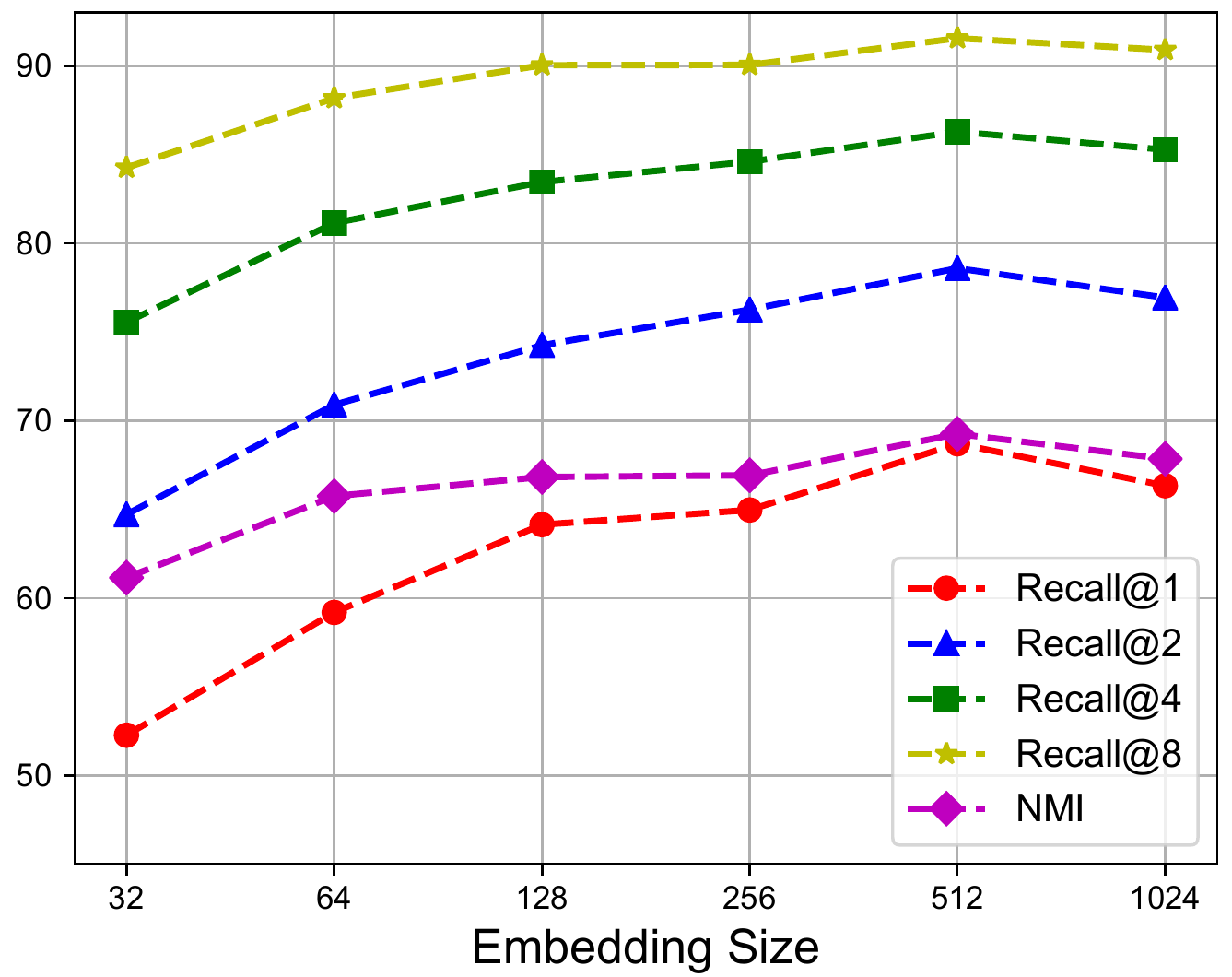}
\caption{Analysis of embedding size.
} 
\label{fig:embedding_size}
\vspace{-5mm}
\end{figure}

\newcommand\figwidth{0.2}

\begin{figure}[tb]
\centering  
\subfigure[Visualization.]{
\label{visualize}
\includegraphics[width=\figwidth\textwidth]{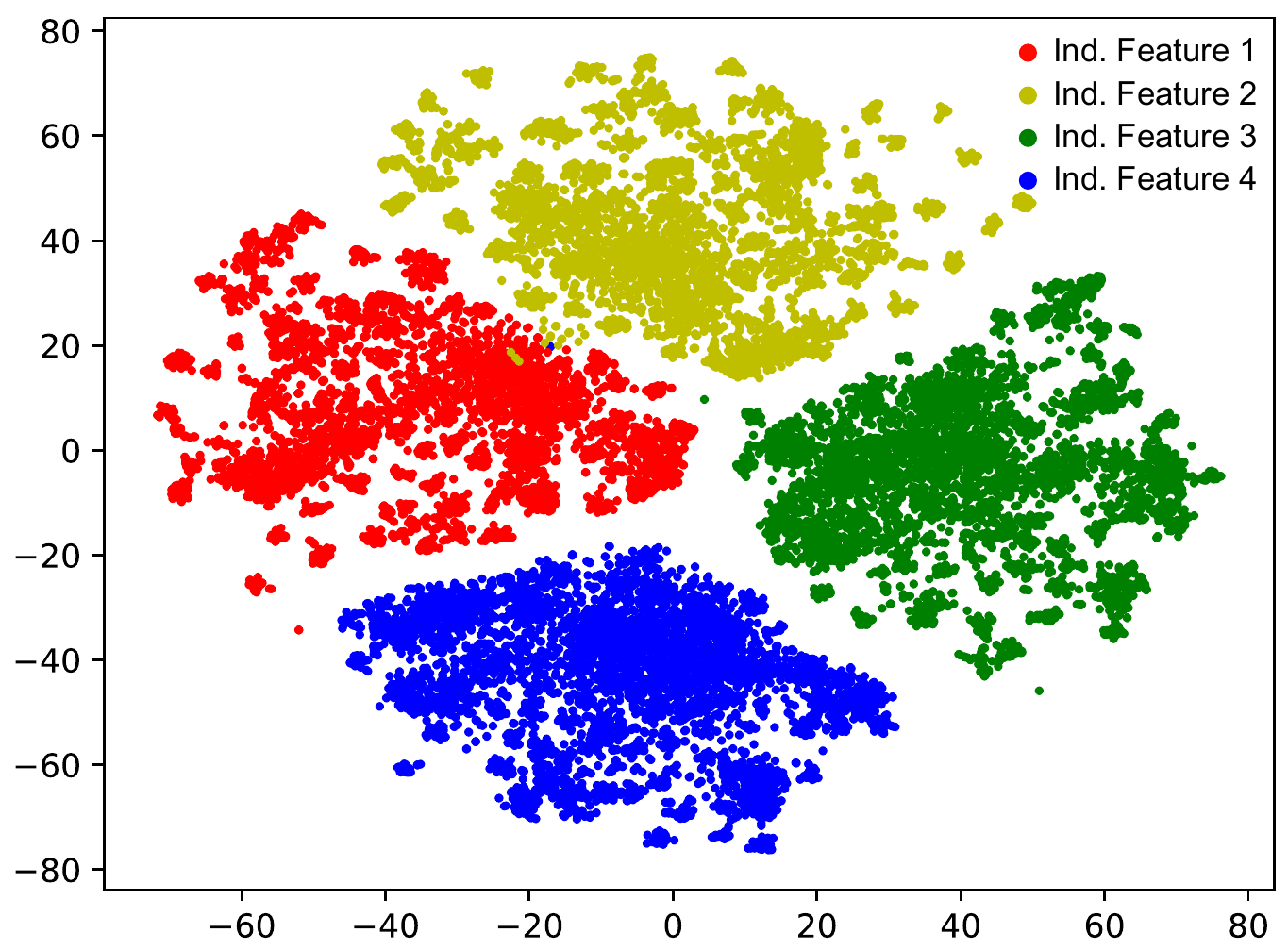}}
\subfigure[Ratio of samples.]{
\label{proportion}
\includegraphics[width=\figwidth\textwidth]{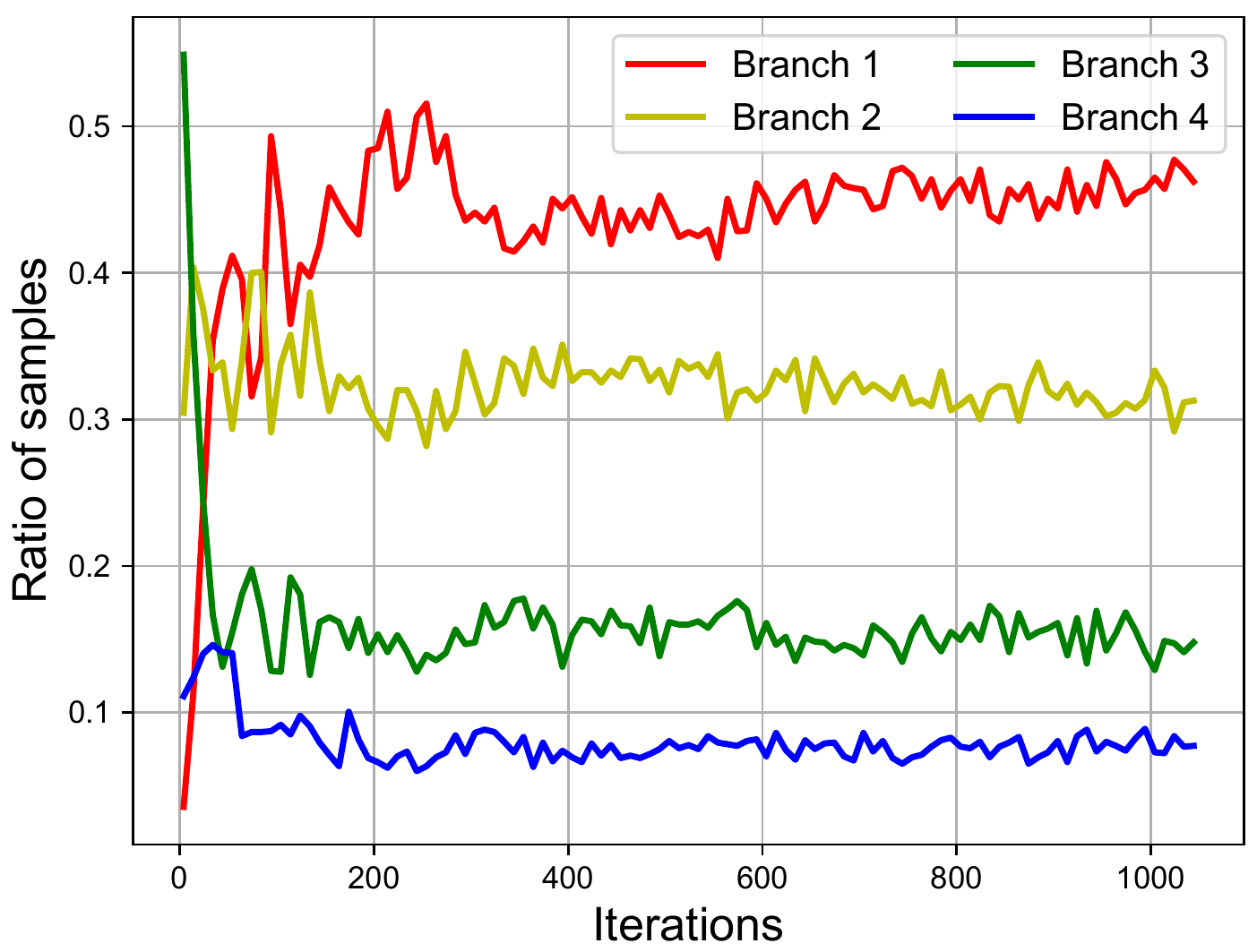}}
\caption{Diversity analysis of the proposed ensemble learning.}
\label{weight}
\vspace{-7mm}
\end{figure}

\textbf{Ablation study of each loss function:}
We also conducted an ablation study of each loss function on the CUB-200-2011 dataset as shown in Table~\ref{ablation_loss}. We see that combining the three proposed loss functions achieves the best result, which demonstrates the effectiveness of each loss.

\textbf{Effectiveness of the learned relations:}
We learn the relations in an end-to-end manner to produce valid embeddings.
To directly verify the effectiveness of the relations, we conducted an experiment to use random relations and fix them while training the other modules with the ProxyAnchor loss on the CUB-200-2011 dataset.
We discover that the R@1 performance decreases from 68.7 to 62.6, which shows that using the learned relations is helpful.

\setlength\tabcolsep{4.5pt}
\begin{table}[t] \tablesize
\centering
\caption{
Experimental results (\%) of DRML in comparison with baseline methods on the CUB-200-2011 dataset. 
}
\label{tab:cub}
\vspace{5pt}
\begin{tabular}{lcccccc}
\toprule  
Methods & Setting & R@1 & R@2 & R@4 & R@8  &NMI \\
\midrule 
N-Pair~\cite{sohn2016improved} & 512G & 50.1 & 63.3 & 74.3 & 83.2 & 60.4 \\
Angular~\cite{wang2017deep} & 512G & 53.6  & 65.0 & 75.3  & 83.7 & 61.0    \\
HDML~\cite{zheng2019hardness} & 512G & 53.7  & 65.7 & 76.7  & 85.7 & 62.6   \\
HTL~\cite{ge2018deep} & 512BN  & 57.1  & 68.8 & 78.7  & 86.5 & -   \\
RLL-H~\cite{wang2019ranked} & 512BN & 57.4 & 69.7 & 79.2 & 86.9 & 63.6 \\
HTG~\cite{zhao2018adversarial}  & 512R & 59.5 & 71.8 & 81.3 & 88.2 & -  \\
Margin~\cite{wu2017sampling} & 128R & 63.6  & 74.4 & 83.1  & 90.0 & 69.0   \\
SoftTriple~\cite{qian2019softtriple} & 512BN & 65.4 & 76.4 & 84.5 & 90.4 & 69.3 \\
Multi-Sim~\cite{wang2019multi} & 512BN & 65.7 & 77.0 & \color{red}86.3 & 91.2 & - \\
MIC~\cite{roth2019mic} & 128R & 66.1 & 76.8 & 85.6 & - & 69.7 \\
DR~\cite{mohan2020moving} & 512BN & 66.1 & 77.0 & 85.1 & 91.1 & - \\
CircleLoss~\cite{sun2020circle} & 512R & 66.7 & 77.4 & \color{blue}86.2 & 91.2 & - \\
RankMI~\cite{kemertas2020rankmi} & 128R & 66.7 & 77.2 & 85.1 & 91.0 & \color{red}71.3 \\
PADS~\cite{roth2020pads} & 128BN & \color{blue}67.3 & \color{blue}78.0 & 85.9 & - & \color{blue}69.9 \\
\midrule
\multicolumn{3}{l}{Ensemble-based methods:} \\
HDC~\cite{yuan2017hard} & 384G & 53.6  & 65.7 & 77.0  & 85.6 & -  \\
A-BIER~\cite{opitz2018deep} & 512G & 57.5 & 68.7 & 78.3 & 86.2 & - \\
ABE-8~\cite{kim2018attention} & 512G & 60.6  & 71.5 & 79.8  & 87.4 & -    \\
Ranked~\cite{wang2019ranked} & 1536BN & 61.3 & 72.7 & 82.7 & 89.4 & 66.1 \\
DREML~\cite{xuan2018deep} & 9216R & 63.9 & 75.0 & 83.1 & 89.7 & 67.8 \\
D \& C~\cite{sanakoyeu2019divide} & 128R & 65.9 & 76.6 & 84.4 & 90.6 & 69.6 \\
\midrule  
Triplet-SH*~\cite{schroff2015facenet}  & 512R & 58.8 & 70.6 & 80.6 & 88.3 & 64.6  \\
DRML-TSH  & 512R & \textbf{61.7} & \textbf{73.5} & \textbf{82.4} & \textbf{88.7} & \textbf{65.6} \\
\midrule  
Margin-DW*~\cite{wu2017sampling}  & 512R & 62.3 & 73.5 & 83.2 & 89.9 & 67.2 \\
DRML-MDW  & 512R & \textbf{65.7} & \textbf{76.9} & \textbf{85.6} & \textbf{91.1} & \textbf{69.0}  \\
\midrule  
ProxyAnchor*~\cite{kim2020proxy}  & 512BN & \color{blue}67.3 & 77.7 & 85.7 & \color{blue}91.4 & 68.7 \\
DRML-PA  & 512BN & \color{red}\textbf{68.7} & \color{red}\textbf{78.6} & \color{red}\textbf{86.3} & \color{red}\textbf{91.6} & \textbf{69.3} \\
\bottomrule
\end{tabular}
\vspace{-7.5mm}
\end{table}

\textbf{Effect of embedding dimension:}
To study the effect of using different embedding dimensions, we conducted a series of experiments on the CUB-200-2011 datasets using an embedding dimension of 32, 64, 128, 256, 512, and 1024.
Figure~\ref{fig:embedding_size} shows the performance of the proposed DRML-PA using different embedding dimensions with a fixed $K=4$. 
We see that the performance improves as the embedding size increases and achieves the best result at 512. 
However, using a larger embedding size (1024) instead hinders the performance, suggesting a redundancy in the embedding space.
Note that our method still maintains a fairly good performance even with a small embedding size like 32, where each individual feature only has a size of 8. 
We think this is because we further exploit the relations among individual features to infer the final embedding, which can partially compensate for the lack of information capacity.

\textbf{Diversity of the ensemble:} 
Since the diversity of the ensemble is important to the performance, we visualized the individual features for all the samples on the CUB-200-2011 dataset using t-SNE in Figure~\ref{visualize}.
We see that the learned individual features are very diverse in the feature space.
This is because we encourage each feature extractor to focus on one set of characteristics by only using the corresponding subset of samples to train it. 
To further study the ensemble training process, we further tracked the ratio of samples assigned to each branch as shown in Figure~\ref{proportion}. 
We observe that all the four feature extractors are properly trained only with different samples, which results in the diversity of the individual features.

\textbf{Quantitative results:}
Since the proposed DRML framework can be generally applied to existing deep metric learning methods, we instantiated our framework with various network architectures, loss functions, and sampling strategies to evaluate the effectiveness, which includes triplet loss with random sampling~\cite{wang2014learning, cheng2016person}.
We applied our framework to the triplet loss with semi-hard sampling (i.e., TSH)~\cite{schroff2015facenet},  the margin loss with distance-weighted sampling (MDW)~\cite{wu2017sampling}, and the ProxyAnchor loss with random sampling (PA)~\cite{kim2020proxy}. 
We use $n$-G/BN/R to indicate the model setting where $n$ is the embedding size and G, BN, R to denote GoogleNet~\cite{szegedy2015going}, BN-Inception~\cite{ioffe2015batch}, and ResNet-50~\cite{he2016deep}, respectively.

\setlength\tabcolsep{4.5pt}
\begin{table}[t] \tablesize
\centering
\caption{
Experimental results (\%) of DRML in comparison with baseline methods on the Cars196 dataset.
}
\label{tab:cars}
\vspace{5pt}
\begin{tabular}{lcccccc}
\toprule 
Methods & Setting & R@1 & R@2 & R@4 & R@8 &NMI \\
\midrule 
N-Pair~\cite{sohn2016improved} & 512G & 71.1 & 79.7 & 86.5 & 91.6 & 64.0 \\
Angular~\cite{wang2017deep} & 512G & 71.3  & 80.7 & 87.0  & 91.8 & 62.4  \\
RLL-H~\cite{wang2019ranked} & 512BN & 74.0 & 83.6 & 90.1 & 94.1 & 65.4 \\
HTG~\cite{zhao2018adversarial}  & 512R & 76.5 & 84.7 & 90.4 & 94.0 & - \\
HDML~\cite{zheng2019hardness} & 512G & 79.1 & 87.1 & 92.1 & 95.5 & 69.7 \\
Margin~\cite{wu2017sampling} & 128R & 79.6 & 86.5 & 91.9  & 95.1 & 69.1  \\
HTL~\cite{ge2018deep} & 512BN & 81.4 & 88.0 & 92.7  & 95.7 & -  \\
MIC~\cite{roth2019mic} & 128R  & 82.6 & 89.1 & 93.2 & - & 68.4\\
RankMI~\cite{kemertas2020rankmi} & 128R & 83.3 & 89.8 & 93.8 & 96.5 & 69.4 \\
CircleLoss~\cite{sun2020circle} & 512R & 83.4 & 89.8 & 94.1 & 96.5 & - \\
PADS~\cite{roth2020pads} & 128BN & 83.5 & 89.7 & 93.8 & - & 68.8 \\
Multi-Sim~\cite{wang2019multi} & 512BN & 84.1 & 90.4 & 94.0 & 96.5 & - \\
SoftTriple~\cite{qian2019softtriple} & 512BN & 84.5 & 90.7 & 94.5 & 96.9 & 70.1 \\
DR~\cite{mohan2020moving} & 512BN & 85.0 & 90.5 & 94.1 & 96.4 & - \\
\midrule
\multicolumn{3}{l}{Ensemble-based methods:} \\
HDC~\cite{yuan2017hard} & 384G & 73.7 & 83.2 & 89.5  & 93.8 & -   \\
A-BIER~\cite{opitz2018deep} & 512G & 82.0 & 89.0 & 93.2 & 96.1 & - \\
Ranked~\cite{wang2019ranked} & 1536BN & 82.1 & 89.3 & 93.7 & 96.7 & 71.8 \\
D \& C~\cite{sanakoyeu2019divide} & 128R & 84.6 &90.7 & 94.1 & 96.5 & 70.3 \\
ABE-8~\cite{kim2018attention} & 512G & 85.2 & 90.5 & 94.0  & 96.1 & -   \\
DREML~\cite{xuan2018deep}  & 9216R & \color{blue}86.0 & \color{blue}91.7 & \color{blue}95.0 & \color{blue}97.2 & \color{red}76.4 \\
\midrule  
Triplet-SH*~\cite{schroff2015facenet}  & 512R & 62.7 & 74.9 & 83.8 & 90.6 & 59.9 \\
DRML-TSH  & 512R & \textbf{64.4} & \textbf{76.4} & \textbf{85.2} &\textbf{91.5} & \textbf{61.0} \\
\midrule  
Margin-DW*~\cite{wu2017sampling}  & 512R & 72.3 & 82.3 & 89.3 & 94.2 & 64.6 \\
DRML-MDW  & 512R & \textbf{73.3} & \textbf{83.0} & \textbf{89.8} & \textbf{94.4} & \textbf{65.3} \\
\midrule  
ProxyAnchor*~\cite{kim2020proxy}  & 512BN & 84.4 & 90.7 & 94.3 & 96.8 & 69.7 \\
DRML-PA  & 512BN & \color{red}\textbf{86.9} & \color{red}\textbf{92.1} & \color{red}\textbf{95.2} & \color{red}\textbf{97.4} & \color{blue}\textbf{72.1} \\
\bottomrule
\end{tabular}
\vspace{-7.5mm}
\end{table}

\setlength\tabcolsep{5.5pt}
\begin{table}[t] \tablesize
\centering
\caption{
Experimental results (\%) of DRML in comparison with baseline methods on the Stanford Online Products dataset. 
}
\label{tab:products}
\vspace{5pt}
\begin{tabular}{lccccc}
\toprule  
Methods & Setting & R@1 & R@10 & R@100 &NMI \\
\midrule 
N-Pair~\cite{sohn2016improved} & 512G & 67.7 & 83.8 & 93.0 & 88.1 \\
Angular~\cite{wang2017deep} & 512G & 67.9 & 83.2 & 92.2 & 87.8   \\
HDML~\cite{zheng2019hardness} & 512G & 68.7 & 83.2 & 92.4 & 89.3   \\
Margin~\cite{wu2017sampling} & 128R & 72.7 & 86.2 & 93.8 & 90.7   \\
RankMI~\cite{kemertas2020rankmi} & 128R & 74.3 & 87.9 & 94.9 & 90.5 \\
HTL~\cite{ge2018deep} & 512BN & 74.8 & 88.3 & 94.8 & -  \\
RLL-H~\cite{wang2019ranked} & 512BN & 76.1 & 89.1 & 95.4 & 89.7 \\
FastAP~\cite{cakir2019deep} & 512R & 76.4 & 89.1 & 95.4 & -   \\
PADS~\cite{roth2020pads} & 128BN & 76.5 & 89.0 & 95.4 & 89.9 \\
MIC~\cite{roth2019mic} & 128R & 77.2 & 89.4 & 95.6 & 90.0 \\
Multi-Sim~\cite{wang2019multi} & 512BN & 78.2 & 90.5 & 96.0 & - \\
SoftTriple~\cite{qian2019softtriple} & 512BN & 78.3 & 90.3 & 95.9 & \color{red}92.0\\
CircleLoss~\cite{sun2020circle} & 512R & 78.3 & 90.5 & \color{blue}96.1 & - \\
\midrule
\multicolumn{3}{l}{Ensemble-based methods:} \\
HDC~\cite{yuan2017hard} & 384G & 70.1 & 84.9 & 93.2 & -   \\
A-BIER~\cite{opitz2018deep} & 512G & 74.2 & 86.9 & 94.0 & -\\
D \& C~\cite{sanakoyeu2019divide} & 128R & 75.9 & 88.4 & 94.9 & 90.2 \\
ABE-8~\cite{kim2018attention} & 512G & 76.3 & 88.4 & 94.8 & -   \\
Ranked~\cite{wang2019ranked} & 1536BN & \color{blue}79.8 & \color{red}91.3 & \color{red}96.3 & 90.4 \\
\midrule  
Triplet-SH*~\cite{schroff2015facenet}  & 512R & 76.1 & 87.7 & 94.3 & 89.3 \\ 
DRML-TSH  & 512R & \textbf{75.7} & \textbf{88.6} & \textbf{95.4} & \textbf{89.8} \\
\midrule  
ProxyAnchor*~\cite{kim2020proxy}  & 512BN & 70.1 & 84.3 & 92.6 & 87.2 \\
DRML-PA  & 512BN & \textbf{71.5} & \textbf{85.2} & \textbf{93.0} & \textbf{88.1} \\
\midrule  
Margin-DW*~\cite{wu2017sampling}  & 512R & 76.8 & 88.9 & 95.1 & 89.7 \\ 
DRML-MDW  & 512R & \color{red}\textbf{79.9} & \color{blue}\textbf{90.7} & \color{blue}\textbf{96.1} & \textbf{90.1} \\
\bottomrule
\end{tabular}
\vspace{-7.2mm}
\end{table}

Tables \ref{tab:cub}, \ref{tab:cars}, and \ref{tab:products} show the experimental results on the CUB-200-2011, Cars196, and Stanford Online Products datasets, respectively, where * denotes our reproduced results under the same setting. We use red numbers to indicate the best results and bold numbers to highlight the improvement of our framework over the associated method without DRML. 
We observe that the proposed framework improves the performance of existing methods on all three datasets. While the use of an advanced sampling scheme can benefit the original loss, the proposed DRML can further boost the performance. 
In particular, our framework performs the best combined with the ProxyAnchor loss and achieves very competitive results. 
Note that the ProxyAnchor loss in the original paper~\cite{kim2020proxy} uses a more sophisticated pooling operation (average + max pooling) and a more advanced optimizer (AdamW) than the conventional metric learning setting, so we only report our reproduced results.
Compared with the original methods, our framework represents the characteristics of an image more comprehensively by learning a feature ensemble with a large diversity and further employs a relational model to capture the structural relations.
 Still, our DRML preserves the advantage of metric learning to learn a discriminative embedding space to effectively measure the similarities among samples.

\vspace{-1mm}
\section{Conclusion}
In this paper, we have presented a deep relational metric learning (DRML) framework for image clustering and retrieval. 
We represent an image by an ensemble of features to capture both interclass and intraclass distributions and employ a relational model to characterize the relations among features. 
We further perform relational inference to produce a relation-aware embedding to measure similarities. 
We have conducted experiments on three datasets which have demonstrated that our method can boost the performance of existing methods on deep metric learning in both image clustering and retrieval tasks. 
It is an interesting future direction to extend our framework to process other formats of input such as video, 3D point clouds, and event-based data.

\vspace{-1mm}
\section*{Acknowledgement}
This work was supported in part by the National Natural Science Foundation of China under Grant U1813218, Grant U1713214, and Grant 61822603, in part by a grant from the Beijing Academy of Artificial Intelligence (BAAI), and in part by a grant from the Institute for Guo Qiang, Tsinghua University.

%{\small
%\bibliographystyle{ieee_fullname}
%\bibliography{egbib}
%}

\clearpage
\appendix
%\setcounter{page}{1}

%\begin{mytitlepage}
%\title{Deep Relational Metric Learning \\ Supplementary Material}
%\twocolumn
%\maketitle
%\input{chapters/8_supplementary.tex}
%\end{mytitlepage}

%\begin{appendices}

\renewcommand{\thesection}{\Alph{section}}
\newcommand{\tablesizesupp}{\tablesize}

\begin{table*}[t] \tablesizesupp
\centering
\caption{Results using the new protocol on the CUB-200-2011 dataset.}
\label{reality_cub}
\begin{tabular}{lccccccccc}
\toprule
\multirow{2}{*}{Method}& 
\multicolumn{3}{c}{Concatenated (512-dim)}&\multicolumn{3}{c}{Separated (128-dim)}  \\
  \cmidrule(lr){2-4} \cmidrule(lr){5-7}   
 & R/P@1 & RP & MAP@R & R/P@1 & RP & MAP@R \\
\midrule
Pretrained & 51.1  & 24.9  & 14.2  & 50.5  & 25.1  & 14.5 \\
Contrastive~\cite{hadsell2006dimensionality} & 67.2 $\pm$ 0.5 & 36.9 $\pm$ 0.3 & 26.2 $\pm$ 0.3 & 58.6 $\pm$ 0.5 & 31.5 $\pm$ 0.2 & 20.7 $\pm$ 0.2 \\
ProxyNCA~\cite{movshovitz2017no} & 66.1 $\pm$ 0.3 & 35.5 $\pm$ 0.2 & 24.6 $\pm$ 0.2 & 58.3 $\pm$ 0.3 & 30.6 $\pm$ 0.1 & 19.7 $\pm$ 0.1 \\
Margin~\cite{wu2017sampling} & 65.5 $\pm$ 0.5 & 35.0 $\pm$ 0.2 & 24.1 $\pm$ 0.3 & 56.2 $\pm$ 0.4 & 29.5 $\pm$ 0.2 & 18.6 $\pm$ 0.2 \\
N. Softmax~\cite{zhai2018classification} & 65.4 $\pm$ 0.2 & 36.0 $\pm$ 0.2 & 25.2 $\pm$ 0.2 & 58.5 $\pm$ 0.2 & 31.7 $\pm$ 0.2 & 20.9 $\pm$ 0.2 \\
ArcFace~\cite{deng2019arcface} & 67.1 $\pm$ 0.3 & 37.2 $\pm$ 0.2 & 26.4 $\pm$ 0.2 & \textcolor[rgb]{ 0,  0,  1}{60.1} $\pm$ 0.2 & \textcolor[rgb]{ 0,  0,  1}{32.3} $\pm$ 0.1 & {21.4} $\pm$ 0.1 \\
FastAP~\cite{cakir2019deep} & 63.6 $\pm$ 0.2 & 34.5 $\pm$ 0.2 & 23.7 $\pm$ 0.2 & 55.9 $\pm$ 0.3 & 29.8 $\pm$ 0.2 & 19.1 $\pm$ 0.2 \\
SNR~\cite{yuan2019signal}   & \textcolor[rgb]{ 0,  0,  1}{67.3} $\pm$ 0.5 & 36.9 $\pm$ 0.2 & 26.1 $\pm$ 0.2 & 58.8 $\pm$ 0.3 & 31.6 $\pm$ 0.2 & 20.8 $\pm$ 0.2 \\
MS~\cite{wang2019multi}    & 66.0 $\pm$ 0.2 & 35.9 $\pm$ 0.1 & 25.2 $\pm$ 0.1 & 58.5 $\pm$ 0.2 & 31.4 $\pm$ 0.1 & 20.6 $\pm$ 0.1 \\
MS+Miner~\cite{wang2019multi} & 65.8 $\pm$ 0.3 & 36.0 $\pm$ 0.2 & 25.2 $\pm$ 0.2 & 58.2 $\pm$ 0.2 & 31.3 $\pm$ 0.2 & 20.5 $\pm$ 0.2 \\
SoftTriple~\cite{qian2019softtriple} & 66.2 $\pm$ 0.4 & 36.5 $\pm$ 0.2 & 25.6 $\pm$ 0.2 & 59.6 $\pm$ 0.4 & 32.1 $\pm$ 0.2 & 21.3 $\pm$ 0.2 \\
\midrule
Triplet~\cite{weinberger2009distance} & \textbf{64.4} $\pm$ 0.4 & 34.6 $\pm$ 0.4 &  23.8 $\pm$ 0.4 & 56.0 $\pm$ 0.3 & 29.6 $\pm$ 0.3 &  18.8 $\pm$ 0.3 \\
DRML-Triplet & 64.2 $\pm$ 0.5  & \textbf{34.8} $\pm$ 0.4 & \textbf{24.1} $\pm$ 0.3 & \textbf{56.3} $\pm$ 0.4 & \textbf{30.0} $\pm$ 0.5 & \textbf{19.3} $\pm$ 0.4 \\
\midrule
ProxyAnchor~\cite{kim2020proxy} & 65.2 $\pm$ 0.2  & 36.0 $\pm$ 0.2 & 25.3 $\pm$ 0.1 & 56.6 $\pm$ 0.1 & 30.5 $\pm$ 0.1 & 19.8 $\pm$ 0.2 \\
DRML-PA & \textbf{66.5} $\pm$ 0.1 & \textbf{36.8} $\pm$ 0.2 & \textbf{26.0} $\pm$ 0.2 & \textbf{59.5} $\pm$ 0.2 & \textbf{32.0} $\pm$ 0.3 & \textbf{21.2} $\pm$ 0.2 \\
\midrule
Cosface~\cite{wang2018cosface} & 67.2 $\pm$ 0.4 & \textcolor[rgb]{ 0,  0,  1}{37.4} $\pm$ 0.2 & \textcolor[rgb]{ 0,  0,  1}{26.5} $\pm$ 0.2 & 59.8 $\pm$ 0.3 & 32.1 $\pm$ 0.2 & \textcolor[rgb]{ 0,  0,  1}{21.6} $\pm$ 0.2 \\
DRML-Cosface & \textcolor[rgb]{ 1,  0,  0}{\textbf{69.2}} $\pm$ 0.3 & \textcolor[rgb]{ 1,  0,  0}{\textbf{37.8}} $\pm$ 0.2 & \textcolor[rgb]{ 1,  0,  0}{\textbf{27.2}} $\pm$ 0.2 & \textcolor[rgb]{ 1,  0,  0}{\textbf{60.2}} $\pm$ 0.3  & \textcolor[rgb]{ 1,  0,  0}{\textbf{33.0}} $\pm$ 0.2 & \textcolor[rgb]{ 1,  0,  0}{\textbf{22.3}} $\pm$ 0.3 \\
\bottomrule
\end{tabular}
%\vspace{-4.5mm}
\end{table*}

\begin{table*}[h] \tablesizesupp
\centering
\caption{Results using the new protocol on the Cars196 dataset.}
\label{reality_cars}
%\vspace{5pt}
\begin{tabular}{lccccccccc}
\toprule
\multirow{2}{*}{Method}& 
\multicolumn{3}{c}{Concatenated (512-dim)}&\multicolumn{3}{c}{Separated (128-dim)}  \\
  \cmidrule(lr){2-4} \cmidrule(lr){5-7}   
 & R/P@1 & RP & MAP@R & R/P@1 & RP & MAP@R \\
\midrule
Pretrained & 46.9  & 13.8  & 5.9   & 43.3  & 13.4  & 5.6  \\
Contrastive~\cite{hadsell2006dimensionality} & 81.6 $\pm$ 0.4 & 35.7 $\pm$ 0.4 & 25.5 $\pm$ 0.4 & 69.4 $\pm$ 0.2 & 28.2 $\pm$ 0.2 & 17.6 $\pm$ 0.2 \\
ProxyNCA~\cite{movshovitz2017no} & 83.3 $\pm$ 0.4 & 36.6 $\pm$ 0.3 & 26.4 $\pm$ 0.4 & 70.9 $\pm$ 0.6 & 28.6 $\pm$ 0.3 & 18.0 $\pm$ 0.3 \\
Margin~\cite{wu2017sampling} & 82.1 $\pm$ 2.4 & 34.7 $\pm$ 2.2 & 24.1 $\pm$ 2.3 & 71.0 $\pm$ 2.7 & 27.6 $\pm$ 1.5 & 16.8 $\pm$ 1.5 \\
N. Softmax~\cite{zhai2018classification} & 83.6 $\pm$ 0.3 & 36.6 $\pm$ 0.2 & 26.4 $\pm$ 0.2 & 72.9 $\pm$ 0.2 & {29.6} $\pm$ 0.1 & {18.9} $\pm$ 0.1 \\
ArcFace~\cite{deng2019arcface} & 84.0 $\pm$ 0.2 & 35.4 $\pm$ 0.3 & 25.2 $\pm$ 0.3 & 73.7 $\pm$ 0.4 & 28.6 $\pm$ 0.1 & 18.1 $\pm$ 0.1 \\
FastAP~\cite{cakir2019deep} & 78.2 $\pm$ 0.7 & 33.4 $\pm$ 0.7 & 22.9 $\pm$ 0.7 & 64.7 $\pm$ 0.6 & 26.4 $\pm$ 0.4 & 15.8 $\pm$ 0.4 \\
SNR~\cite{yuan2019signal} & 81.9 $\pm$ 0.4 & 35.4 $\pm$ 0.4 & 25.1 $\pm$ 0.5 & 70.2 $\pm$ 0.4 & 27.9 $\pm$ 0.4 & 17.4 $\pm$ 0.3 \\
MS~\cite{wang2019multi} & {85.3} $\pm$ 0.3 & \textcolor[rgb]{ 0,  0,  1}{38.0} $\pm$ 0.6 & \textcolor[rgb]{ 0,  0,  1}{27.8} $\pm$ 0.8 & 73.7 $\pm$ 1.0 & 29.4 $\pm$ 0.6 & 18.8 $\pm$ 0.7 \\
MS+Miner~\cite{wang2019multi} & 84.6 $\pm$ 0.3 & 37.7 $\pm$ 0.4 & 27.6 $\pm$ 0.4 & 72.9 $\pm$ 0.3 & 29.5 $\pm$ 0.4 & 18.9 $\pm$ 0.4 \\
SoftTriple~\cite{qian2019softtriple} & 83.7 $\pm$ 0.2 & 36.3 $\pm$ 0.2 & 26.1 $\pm$ 0.2 & 73.0 $\pm$ 0.2 & 29.4 $\pm$ 0.1 & 18.7 $\pm$ 0.1 \\
\midrule
Triplet~\cite{weinberger2009distance} & 77.5  $\pm$ 0.6 & 32.9 $\pm$ 0.5 & 22.1 $\pm$ 0.5  & \multicolumn{1}{l}{63.9 $\pm$ 0.4} & 26.1 $\pm$ 0.3 & 15.2 $\pm$ 0.3 \\
DRML-Triplet & \textbf{78.8} $\pm$ 0.3 & \textbf{33.2} $\pm$ 0.4 & \textbf{22.8} $\pm$ 0.5 & \textbf{64.0} $\pm$ 0.2 & \textbf{26.2} $\pm$ 0.3 & \textbf{15.5} $\pm$ 0.1 \\
\midrule
ProxyAnchor~\cite{kim2020proxy} & 83.3 $\pm$ 0.4  & 35.7 $\pm$ 0.3  & 25.7 $\pm$ 0.4 & 73.7 $\pm$ 0.4 & 29.4 $\pm$ 0.3 & 18.9 $\pm$ 0.2 \\
DRML-PA & \textcolor[rgb]{ 0,  0,  1}{\textbf{85.7}} $\pm$ 0.5 & \textbf{36.0} $\pm$ 0.2 & \textbf{26.1} $\pm$ 0.2 & \textcolor[rgb]{ 1,  0,  0}{\textbf{76.6}} $\pm$ 0.4 & \textcolor[rgb]{ 0,  0,  1}{\textbf{29.8}} $\pm$ 0.3 & \textcolor[rgb]{ 0,  0,  1}{\textbf{19.3}} $\pm$ 0.2 \\
\midrule
Cosface~\cite{wang2018cosface} & 85.3 $\pm$ 0.2 & 36.7 $\pm$ 0.2 &  26.9 $\pm$ 0.2 & {74.1} $\pm$ 0.2 & 28.5 $\pm$ 0.1 &  18.2 $\pm$ 0.1 \\
DRML-Cosface & \textcolor[rgb]{ 1,  0,  0}{\textbf{86.4}} $\pm$ 0.3 & \textcolor[rgb]{ 1,  0,  0}{\textbf{38.7 }} $\pm$ 0.4 & \textcolor[rgb]{ 1,  0,  0}{\textbf{29.2}} $\pm$ 0.3 & \textcolor[rgb]{ 0,  0,  1}{\textbf{75.7}} $\pm$ 0.3 & \textcolor[rgb]{ 1,  0,  0}{\textbf{30.2}} $\pm$ 0.2 & \textcolor[rgb]{ 1,  0,  0}{\textbf{20.0}} $\pm$ 0.1 \\
\bottomrule
\end{tabular}
%\vspace{-5mm}
\end{table*}

\setlength\tabcolsep{7pt}
\vspace{-3.5mm}
\begin{table*}[h] \tablesizesupp
\centering
\caption{Experimental results on the ImageNet dataset.}
\label{imagenet}
\begin{tabular}{lcccccccc}
\toprule
Method & R/P@1 & R@2 & P@2 & RP  & MAP@R & NMI \\
\midrule 
Softmax Baseline & 53.7  & 63.8  & 50.9  & 25.5  & 33.9  & 71.8  \\
\midrule
Margin-DW~\cite{wu2017sampling} & 46.3  & 56.5  & 45.5  & 23.7  & 33.1  & 74.6  \\
DRML-MDW & \textbf{48.9 } & \textbf{59.2 } & \textbf{48.3 } & \textbf{25.1 } & \textbf{34.4 } & \textbf{75.4 } \\
\midrule
Triplet-SH*~\cite{schroff2015facenet} & 54.9  & 64.5  & 55.2  & 32.0  & 41.3  & 78.3  \\
DRML-TSH & \textbf{55.8 } & \textbf{65.3 } & \textbf{55.3 } & \textbf{32.3 } & \textbf{41.5 } & \textbf{78.6 } \\
\midrule
ProxyAnchor~\cite{kim2020proxy} & 66.4  & 74.1  & 66.4  & 44.2  & 52.2  & 82.2  \\
DRML-PA & \textcolor[rgb]{ 1,  0,  0}{\textbf{68.0 }} & \textcolor[rgb]{ 1,  0,  0}{\textbf{75.0 }} & \textcolor[rgb]{ 1,  0,  0}{\textbf{67.6 }} & \textcolor[rgb]{ 1,  0,  0}{\textbf{46.3 }} & \textcolor[rgb]{ 1,  0,  0}{\textbf{53.9 }} & \textcolor[rgb]{ 1,  0,  0}{\textbf{82.9 }} \\
\bottomrule
\end{tabular}
%\vspace{-3.5mm}
\end{table*}

\begin{figure*}[h]
\centering
\includegraphics[width=1\textwidth]{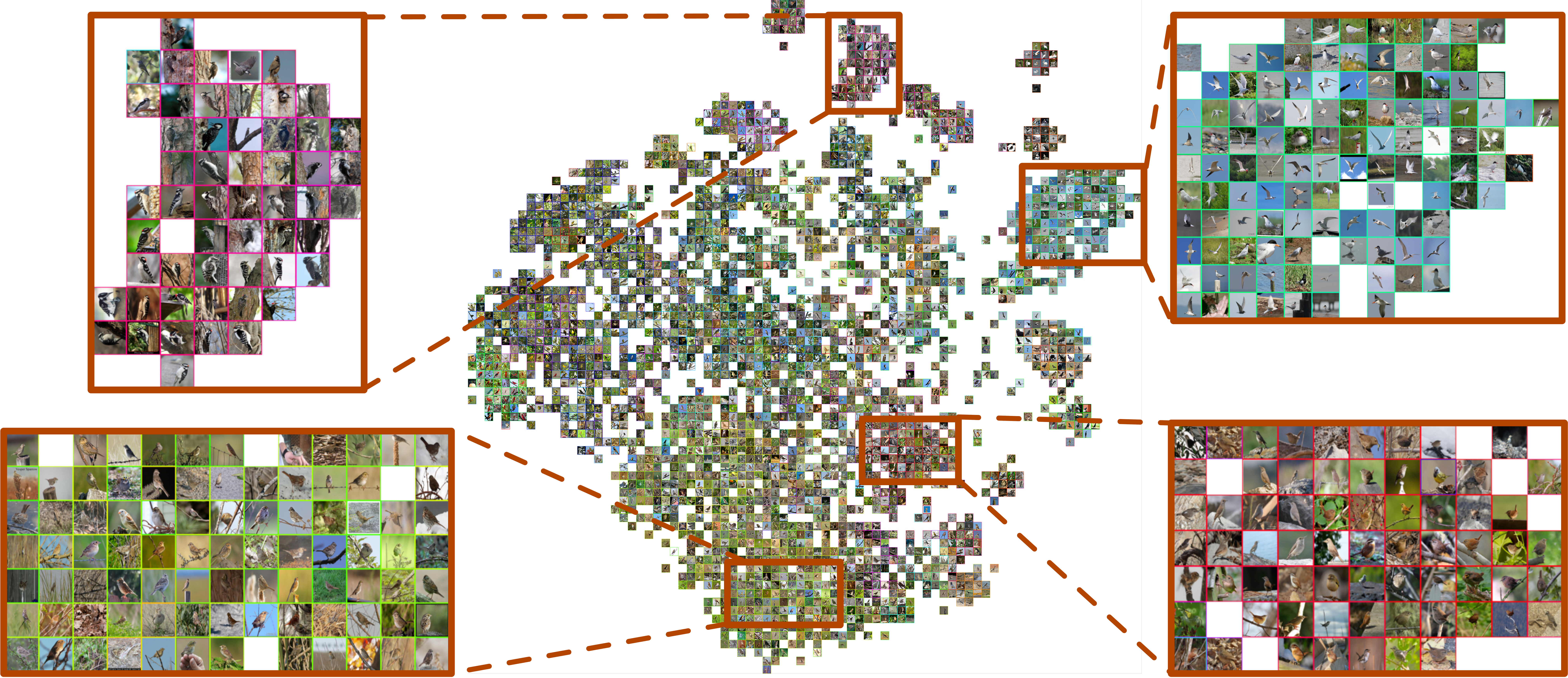}
\caption{
Qualitative result of the proposed DRML-MDW method on the test subset of the CUB-200-2011 dataset, where we magnify specific regions for clear demonstration.
(Best viewed on a monitor when zoomed in.)} 
\label{tsne_cub}
%\vspace{-4mm}
\end{figure*}

\section{Results using the evaluation protocol~\cite{musgrave2020metric}}
Though we followed the standard evaluation protocol~\cite{song2016deep,zheng2019hardness,sohn2016improved} and used a constrained experimental setting for fair comparisons with existing deep metric learning methods, the conclusions can still be questioned due to the lack of a validation set and the uninformative evaluation metric~\cite{musgrave2020metric}.
To improve the credibility of our experimental evaluation, we additionally performed experiments on the CUB-200-2011~\cite{wah2011caltech} and Cars196~\cite{krause20133d} dataset by strictly following the new evaluation protocol~\cite{musgrave2020metric}.

Specifically, we employed a BN-Inception~\cite{ioffe2015batch} network pretrained on ImageNet~\cite{russakovsky2015imagenet} as the trunk model. 
We set the dimension of the final embedding to 128 and use a batch size of 32 for training.
To prevent direct test set feedback, we performed a 4-fold cross-validation on the training subset to search for the hyperparameters.
We used the first half of the classes as the training subset and the rest as the test subset, and then evenly split the training subset into four partitions based on the number of classes.
During each validation, we employed one of the four partitions for training the rest for evaluation. 
We used the average accuracy on the four validation sets as feedback to tune the hyperparameters.

For testing, we reported the performance in separated and concatenated setting.
For the separated setting, we directly computed the performance of the four 128-dim embeddings obtained using the model trained in each fold and reported the average results.
For the concatenated setting, we concatenated the four aforementioned embedding for each sample to obtain a 512-dim embeddings for evaluation.
We employed the Precision@1 (R/P@1), the R-Precision (RP), and the Mean Average Precision at R (MAP@R) as the evaluation metric.
We direct interesting readers to the original paper~\cite{musgrave2020metric} for more details.

Table~\ref{reality_cub} and \ref{reality_cars} shows the results of the baseline methods and the proposed DRML framework on the CUB-200-2011 and Cars196 dataset, respectively.
We use red numbers to denote the best results and blue numbers to denote the second best results.
We applied our framework to the triplet loss~\cite{weinberger2009distance}, the ProxyAnchor loss~\cite{kim2020proxy}, and Cosface~\cite{wang2018cosface}.
We see that our DRML framework still consistently boosts the performance of existing methods and further achieves the state-of-the-art result under the new evaluation protocol, which verifies the effectiveness of the proposed relation-aware embedding.

\section{Performance on large-scale datasets}
We further conducted experiments on the ImageNet dataset~\cite{russakovsky2015imagenet} to evaluate the generalization of the proposed method to large-scale datasets.
Table~\ref{imagenet} shows the results of our DRML framework applied existing deep metric learning methods.
As the original papers did not reported the performance on the ImageNet dataset, the results in Table~\ref{imagenet} are based on our reproduction\footnote{Code: \url{https://github.com/zbr17/DRML}}.
We observe that the ProxyAnchor loss with random sampling (PA)~\cite{kim2020proxy} is the best baseline method.
The triplet loss with semi-hard sampling (i.e., TSH)~\cite{schroff2015facenet} achieves better results than the softmax baseline while the margin loss with distance-weighted sampling (MDW)~\cite{wu2017sampling} achieves worse results, though MDW consistently outperforms TSH on small-scale datasets like the CUB-200-2011 and Cars196 datasets. 
We can see this trend on the middle-scale Stanford Online Products~\cite{song2016deep} dataset as the two methods achieve comparable performance.
Despite the changing ranking of performance on datasets of different scales, our DRML framework can uniformly improve the performance of various methods, which shows that the effectiveness of our framework generalizes well to the benchmark scale.

\section{Visualization of the embedding space}

Figure \ref{tsne_cub} shows the qualitative result of the proposed DRML-MDW on the CUB-200-2011 dataset.
We employed the Barnes-Hut t-SNE~\cite{vandermaaten2014accelerating} algorithm to visualize the learned embedding space and magnify specific regions for clear demonstration. We color the boundary of each image using different colors to represent the ground truth class label. We observe that even though the classes in the test subset are not seen during training, our method can still accurately measure their semantic differences. Moreover, the images in the CUB-200-2011 dataset possess small interclass differences and large intraclass variations, 
yet our framework still effectively clusters together instances from the same class using the learned relation-aware embeddings despite all these difficulties.

%\end{appendices}

{\small
\bibliographystyle{ieee_fullname}

}

\end{document}